\documentclass[manuscript,screen]{acmart}

\AtBeginDocument{%
  \providecommand\BibTeX{{%
    \normalfont B\kern-0.5em{\scshape i\kern-0.25em b}\kern-0.8em\TeX}}}

\copyrightyear{2021}
\acmYear{2021}
\setcopyright{licensedothergov}\acmConference[DASIP '21]{Workshop on Design and Architectures for Signal and Image Processing (14 th edition)}{January 18--20, 2021}{Budapest (initially), Hungary}
\acmBooktitle{Workshop on Design and Architectures for Signal and Image Processing (14 th edition) (DASIP '21), January 18--20, 2021, Budapest (initially), Hungary}
\acmPrice{15.00}
\acmDOI{10.1145/3441110.3441575}
\acmISBN{978-1-4503-8901-3/21/01}

\usepackage[acronym,toc]{glossaries}
\glsdisablehyper

\newacronym{ale}{ALE}{Arcade Learning Environment}
\newacronym{ai}{AI}{Artificial Intelligence}

\newacronym{dnn}{DNN}{Deep Neural Network}

\newacronym{gegelati}{\textsc{Gegelati}}{Generic Evolvable Graphs for Efficient Learning of Artificial Tangled Intelligence}
\newacronym{gpu}{GPU}{Graphics Processing Unit}

\newacronym{iot}{IoT}{Internet of Things}

\newacronym[longplural=Multiprocessor Systems-on-Chips]{mpsoc}{MPSoC}{Multiprocessor System-on-Chip}

\newacronym{nlp}{NLP}{Natural Language Processing}

\newacronym{pe}{PE}{Processing Element}
\newacronym{prng}{PRNG}{Pseudo-Random Number Generator}

\newacronym{tpg}{TPG}{Tangled Program Graph}

\usepackage{wrapfig}

\usepackage[ruled,linesnumbered,commentsnumbered,procnumbered]{algorithm2e}
\SetNoFillComment 
\SetKwFor{ForEach}{for each}{do}{endfor}

\usepackage{pgf,tikz}
\usetikzlibrary{arrows}
\usetikzlibrary{calc}
\usetikzlibrary{arrows.meta}

\DeclareTextFontCommand{\labelFont}{\fontfamily{phv}\fontsize{8pt}{7pt}\selectfont} 

\DeclareTextFontCommand{\smallFont}{\fontfamily{phv}\fontsize{5pt}{4pt}\selectfont} 

\usepackage{subcaption}

\usepackage{graphics}

\usepackage{bm}

\usepackage{listings}
\lstdefinestyle{customcpp}{
	belowcaptionskip=1\baselineskip,
	breaklines=true,
	frame=single,
	language=C++,
	showstringspaces=false,
	basicstyle=\footnotesize\ttfamily,
	keywordstyle=\bfseries\color{green!40!black},
	commentstyle=\itshape\color{purple!40!black},
	identifierstyle=\color{blue},
	stringstyle=\color{orange},
	captionpos=b
}

\usepackage{lipsum}

\begin{document}

\title[\textsc{Gegelati}: Lightweight AI through Generic and Evolvable TPGs]{\textsc{Gegelati}: Lightweight Artificial Intelligence through Generic and Evolvable Tangled Program Graphs}

\ifx\anonymous\undefined
\author{K. Desnos}
\email{kdesnos@insa-rennes.fr}
\orcid{0000-0003-1527-9668}
\author{N. Sourbier}
\email{nsourbie@insa-rennes.fr}
\author{P.-Y. Raumer}
\email{plerolla@insa-rennes.fr}
\affiliation{%
  \institution{Univ Rennes, INSA Rennes, CNRS, IETR - UMR6164}
  \city{Rennes}
  \country{France}
}

\author{O. Gesny}
\email{ogesny@silicom.fr}
\affiliation{%
	\institution{Silicom}
	\city{Rennes}
	\country{France}
}

\author{M. Pelcat}
\email{mpelcat@insa-rennes.fr}
\affiliation{%
	\institution{Univ Rennes, INSA Rennes, CNRS, IETR - UMR6164}
	\city{Rennes}
	\country{France}
}

\renewcommand{\shortauthors}{Desnos et al.}
\else
\author{Link}
\email{name@email.com}
\affiliation{%
	\institution{Hyrule Kingdom}
}

\author{Mario}
\email{name@email.com}
\affiliation{%
	\institution{Mushroom Kingdom}
}

\author{Ash}
\email{name@email.com}
\affiliation{%
	\institution{Pallet Town}
}

\renewcommand{\shortauthors}{Link et al.}

\fi

\begin{abstract}
  %
  \gls{tpg} is a reinforcement learning technique based on genetic programming concepts.
  On state-of-the-art learning environments, \glspl{tpg} have been shown to offer comparable competence with \glspl{dnn}, for a fraction of their computational and storage cost.
  This lightness of \glspl{tpg}, both for training and inference, makes them an interesting model to implement \glspl{ai} on embedded systems with limited computational and storage resources.

  In this paper, we introduce the \acrshort{gegelati} library for \glspl{tpg}.
  Besides introducing the general concepts and features of the library, two main contributions are detailed in the paper: 
  1/ The parallelization of the deterministic training process of \glspl{tpg}, for supporting heterogeneous \glspl{mpsoc}.
  2/ The support for customizable instruction sets and data types within the genetically evolved programs of the \gls{tpg} model.
  %
  The scalability of the parallel training process is demonstrated through experiments on architectures ranging from a high-end 24-core processor to a low-power heterogeneous \gls{mpsoc}.
  The impact of customizable instructions on the outcome of a training process is demonstrated on a state-of-the-art reinforcement learning environment.
\end{abstract}


\maketitle

\glsresetall

\section{Introduction}
\label{sec:introduction}
 
In less than a decade, \glspl{ai} powered by \glspl{dnn} have outperformed and replaced man-made algorithms in many applicative domains, from computer vision~\cite{Canziani_analysis_2016} to \gls{nlp}~\cite{Brown_Language_2020}.
This sudden breakthrough of \glspl{dnn} is largely due to the availability of affordable and easily programmable hardware offering important computing power, such as \glspl{gpu}~\cite{Krizhevsky_ImageNet_2012}.
Powered by the ever-increasing computing power of commercial chips, the current race for omnipotent \glspl{ai} leads to the creation of more and more complex \glspl{dnn} where millions~\cite{Canziani_analysis_2016} to hundreds of billions~\cite{Brown_Language_2020} parameters are needed.

Conversely to the increasing computational complexity of \glspl{dnn}, the need for lightweight \glspl{ai} is also growing.
Indeed, the ubiquity of \gls{iot} devices, and the tremendous amount of data they generate~\cite{CISCO_Cisco_2018}, call for new paradigms where data processing is performed on-site, close to the data producer.
The processing of data by \glspl{ai} on embedded \gls{iot} devices is not compatible with the use of compute, memory and power-hungry \glspl{dnn}.
A common way to create lightweight \glspl{ai} is to exploit the resilience of \glspl{dnn} to approximation, and to simplify them as much as possible while maintaining their accuracy to acceptable levels.
Pruning techniques, customizable data precision, and approximate computing techniques are examples of techniques for reducing \glspl{dnn} complexity.
An alternative way to create lightweight \glspl{ai} is to develop new machine learning techniques that rely on light-by-construction models, such as the \acrshort{tpg} model studied in this paper.

\acrshort{tpg}, which stands for \acrlong{tpg}, is a machine learning model proposed by Kelly and Heywood in~\cite{Kelly_Emergent_2017}.
Building on state-of-the-art genetic programming techniques, \glspl{tpg} are grown from scratch for each learning environment in which they are trained.
Hence, the topology and the complexity of the \gls{tpg} adapt themselves to the complexity of the learned task, without requiring an expert to select an appropriate network structure.
In recent works~\cite{Kelly_Emergent_2017, Kelly_Modular_2020, Kelly_Scaling_2018}, \glspl{tpg} have proven to be a very promising model for building \glspl{ai}, being competitive with state-of-the-art \glspl{dnn} for a fraction of their computation and storage cost, both for training and inference.

This paper introduces \acrshort{gegelati}\footnote{\gls{gegelati}}, an open-source library for the learning and inference of \glspl{ai} modeled with \glspl{tpg}.
The objective of \gls{gegelati}, coded in C++, is to foster the development of efficient, lightweight and portable \glspl{ai}, supporting both general-purpose and embedded hardware.
The modular structure of the library fosters its extensibility and customizability to ease its evolutions and its adaptations for new learning tasks.

The principles of \gls{tpg}-based \glspl{ai} and related works are presented in Section~\ref{sec:tpg}.
Section~\ref{sec:gegelati} presents the \gls{gegelati} library.
Two distinctive features of \gls{gegelati} are detailed in this section. 
First, the parallel, scalable, and yet deterministic training of \glspl{tpg} on heterogeneous multicore architectures is introduced in  Section~\ref{sec:parallelism}.
Second, the support for customizable instruction sets for easing the training of \glspl{tpg} in diverse learning environments is detailed in Section~\ref{sec:custom_instructions}.
Experiments on various learning environments are presented in Section~\ref{sec:experiments}, demonstrating the scalable performance and the customizability of the library.
Finally, Section~\ref{sec:conclusion} concludes this paper.

\section{Tangled Program Graphs}
\label{sec:tpg}

The \gls{tpg} model studied in this paper, which builds on technique from the genetic programming domain, was introduced by Kelly and Heywood~\cite{Kelly_Emergent_2017} as a reinforcement learning technique.
Principles of reinforcement learning and genetic programming are presented in Section~\ref{sec:reinforcement_learning}, and the \gls{tpg} model is detailed in Section~\ref{sec:tpg_model}.

\subsection{Background: Reinforcement Learning and Genetic Programming}
\label{sec:reinforcement_learning}

\textbf{Reinforcement learning} is a branch of machine learning techniques where artificial intelligence learns, through trial and error, how to interact with an environment.
In reinforcement learning, artificial intelligence, called the learning agent, observes the current state of its learning environment, and interacts with it trough a finite set of actions. 
As a result of these actions, or because of external phenomena such as time or physics, the state of the learning environment evolves.
By observing the constantly evolving state of the environment, the learning agent has the possibility to react and to build a meaningful sequence of actions.
For the agent to learn which sequences of actions are useful, an additional reward mechanism is implemented.
By rewarding useful behavior of the learning agent, and penalizing harmful or useless behavior, this reward mechanism helps the learning agent select the most appropriate behavior for each new experience.
Although \glspl{tpg} have originally been developed for reinforcement learning purposes, the possibility to adapt them for other kinds of learning environments has already been demonstrated~\cite{Kelly_Modular_2020}.

\textbf{Genetic programming} is a subset of machine learning techniques that mimics the natural selection evolution process to breed programs for a selected purpose.
The iterative learning process of genetic programming can be summarized in four steps: 1/ Create an initial population of $n\in \mathbb{N}^*$ random programs.
Then, iteratively: 
2/ Evaluate the fitness of these programs against the learning environment. 
3/ Discard the $m<n, m\in\mathbb{N}^*$ programs of the population with the worse fitnesses.
4/ Recreate $m$ new programs from remaining programs by using genetic operations, like mutations or crossovers.  
As detailed in~\cite{Kelly_Emergent_2017, Kelly_Scaling_2018}, \glspl{tpg} add a compositional mechanism to this genetic learning process, which favors the emergence of stable clusters of useful programs by building a hierarchical decision structure.

\subsection{TPG: Model and Learning Algorithm}
\label{sec:tpg_model}
\begin{figure}[t]
	\centering
	\subcaptionbox{\gls{tpg} example}{
		\includegraphics[scale=0.5]{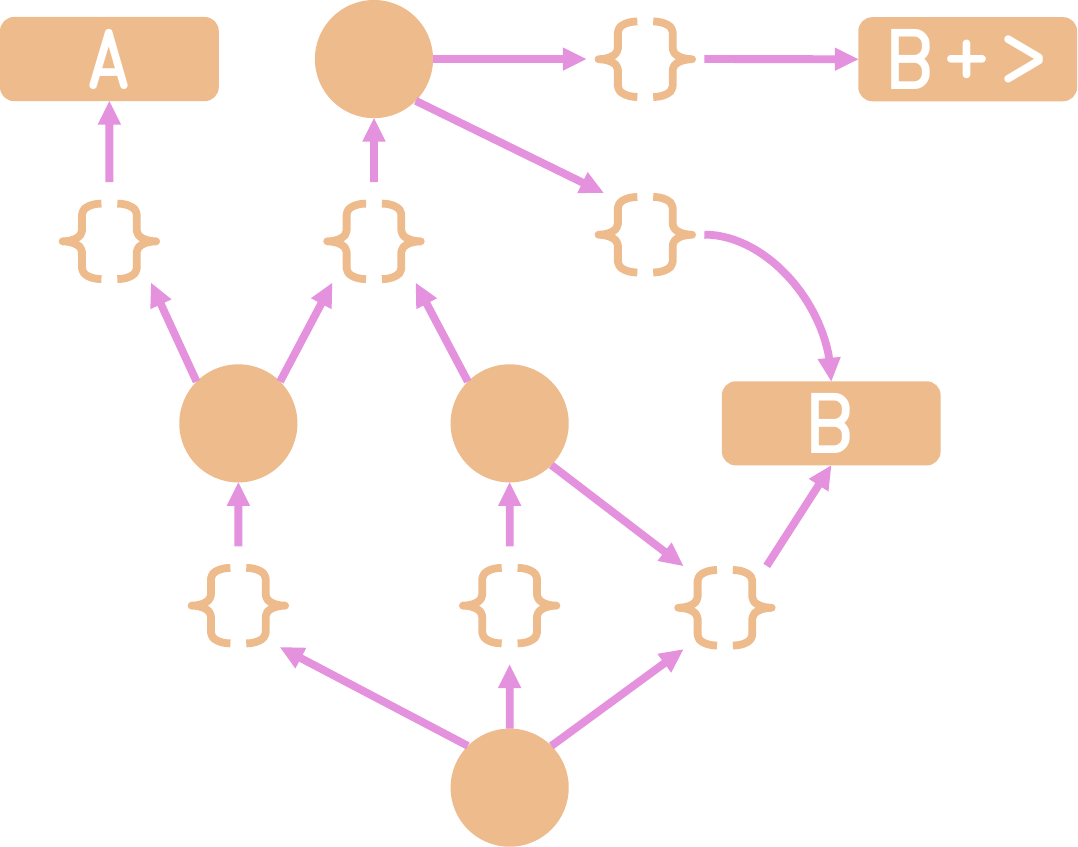}		
		\label{fig:tpg_example}
	}
	\hspace{3em}
	\subcaptionbox{\gls{tpg} semantics}{
		\includegraphics[scale=0.5]{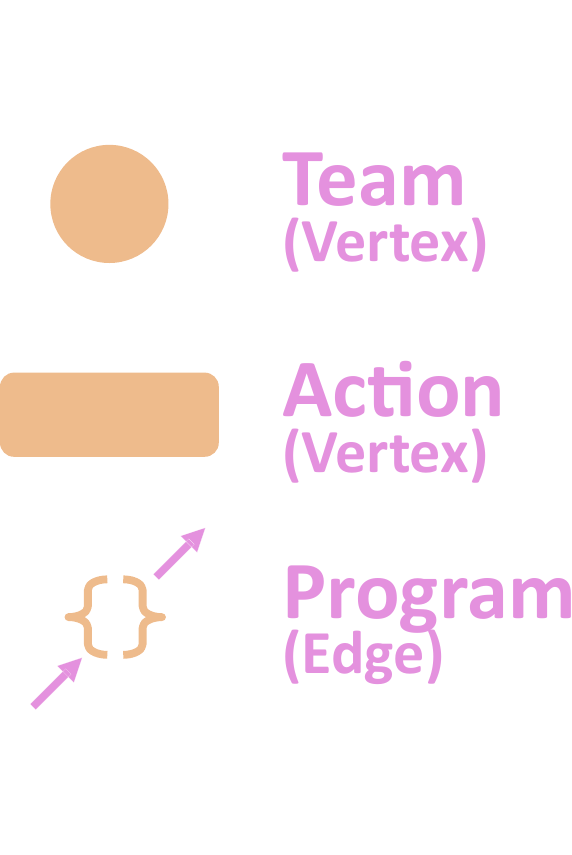} 
		\label{fig:tpg_semantics}
	}
	\caption{Semantics of the \acrfullpl{tpg}}
	\label{fig:tpg}
	\vspace{-1.5em}
\end{figure}
\textbf{The semantics of the \acrfull{tpg} model,} depicted in Figure~\ref{fig:tpg}, consists of three elements composing a direct graph: \textit{programs}, \textit{teams} and \textit{actions}.
The \textit{teams} and the \textit{actions} are the vertices of the graph, \textit{teams} being internal vertices, and \textit{actions} being the leaves of the graph. 
The \textit{programs}, associated to the edges of the graph that each connects a source \textit{team} to a destination \textit{team} or \textit{action} vertex.
Self-loops, that is an edge connecting a \textit{team} to itself, are not allowed in \glspl{tpg}.

\setlength{\intextsep}{0pt}
\begin{wrapfigure}[10]{r}{8cm}
	\includegraphics[scale=0.37]{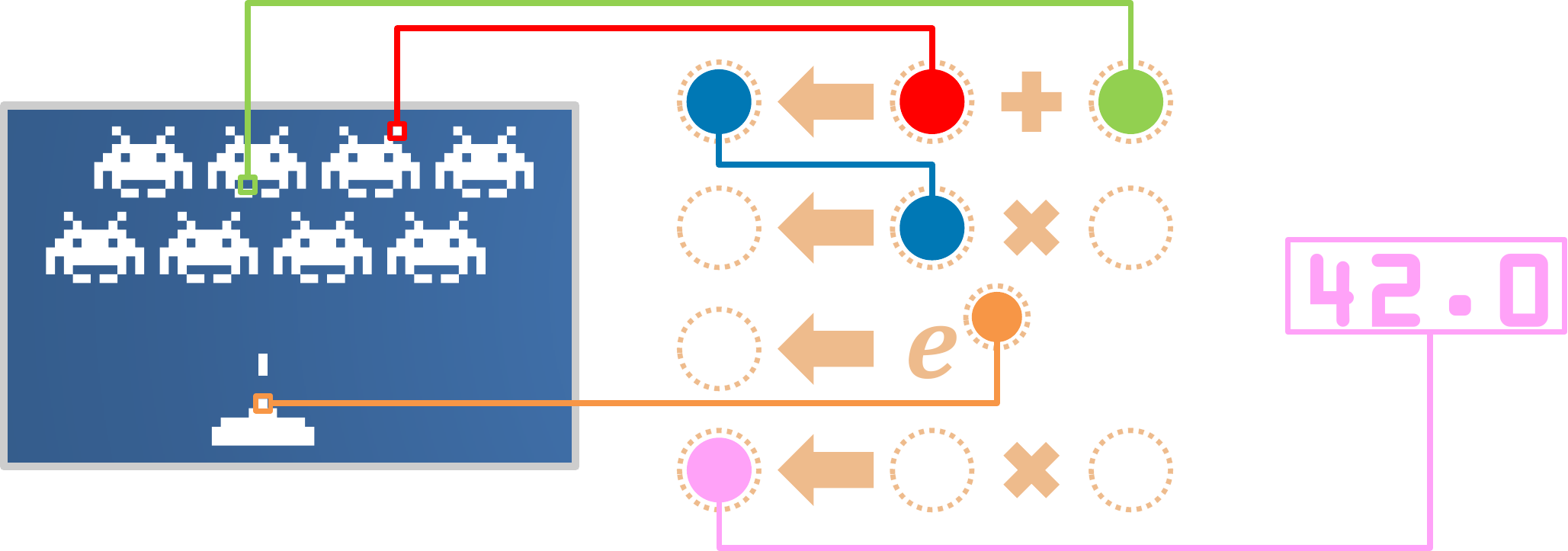}
	\caption{\textit{Program} from a \gls{tpg}. On the left, the learning environment state fed to the \textit{program}. In the middle, the sequence of instructions of the \textit{program}. On the right, the result produced by the program.}
	\label{fig:program}
\end{wrapfigure}
From afar, a \textit{program} can be seen as a black box that takes the current state of the learning environment as an input, processes it, and produces a real number, called a \textit{bid}, as a result.
In more detail, a \textit{program} is a sequence of simple arithmetic \textit{instructions}, like additions or exponents.
As depicted in Figure~\ref{fig:program}, each \textit{instruction} takes as an operand either data coming from the observed learning environment, or the value stored in a register by a previous \textit{instruction}.
The last value stored in a specific register, generally called \texttt{R0}, is the result produced by the \textit{program}.

\textbf{The execution of a \gls{tpg}} starts from its unique root \textit{team}, when a new state of the environment becomes available.
All \textit{programs} associated to outgoing edges of the root \textit{team} are executed with the current state of the environment as their input.
Once all \textit{programs} have completed their execution, the edge associated to the largest \textit{bid} is identified, and the execution of the \gls{tpg} continues following this edge.
If another \textit{team} is pointed by this edge, its outgoing \textit{programs} are executed, still with the same input state, and the execution continues along the edge with the largest \textit{bid}\footnote{If a team is visited several times, previously taken edges are ignored to avoid infinite loops.}.
Eventually, the edge with the largest \textit{bid} leads to an \textit{action} vertex.
In this case, the \textit{action} is executed by the learning agent, a new resulting state of the environment is received, and the \gls{tpg} execution restarts from its root \textit{team}.

\textbf{The genetic evolution process of a \gls{tpg}} relies on a graph with several root \textit{teams}.
The initial \gls{tpg} created for the first generation only contains root \textit{teams} whose outgoing edges each lead directly to an \textit{action} vertex.
At a given generation of the learning process, each root \textit{team} of the \gls{tpg} represents a different \textit{policy} whose fitness is evaluated.
Evaluating a root \textit{team} consists of executing the \gls{tpg} stemming from it a fixed number of times, or until a terminal state of the learning environment is reached, like a game-over in a video game.
The rewards obtained after evaluating each root \textit{team} of the \gls{tpg} are used by the genetic evolution process.
Worst-fitting root \textit{teams}, which obtained the lowest rewards, are deleted from the \gls{tpg}.

To create new root \textit{teams} for the next generation of the evolution process, randomly selected remaining \textit{teams} from the \gls{tpg} are duplicated with all their outgoing edges.
Then, these new edges undergo a random mutation process, possibly altering their destination vertex, and modifying their \textit{programs} by adding, removing, swapping, and changing their instructions and operands.
Surviving root \textit{teams} from previous generations may become the destination of an edge added during the mutation process, thus becoming internal vertices of the \gls{tpg}.
This mutation mechanism favors the emergence of long-living valuable subgraphs of connected \textit{teams}.
Indeed, useful \textit{teams} contributing to higher rewards have a greater chance of becoming internal vertices of the \gls{tpg} which can not be discarded unless they become root \textit{teams} again.
Hence, complexity is added to the \gls{tpg} adaptively, only if this complexity leads to better rewards for the learning agent.
A detailed description of this evolution process can be found in~\cite{Kelly_Scaling_2018}.

\textbf{The capabilities of \glspl{tpg}} have been extensively demonstrated~\cite{Kelly_Emergent_2017, Kelly_Scaling_2018} on the 55 video games from the \gls{ale}~\cite{Bellemare_Arcade_2013}.
In this learning environment, the adaptive complexity leads to \gls{tpg} with diverse sizes, depending on the complexity of the strategies developed to play each game. 
For example, there are two orders of magnitude between the smallest and largest networks built within these learning environments.
On the performance side, \glspl{tpg} have been shown to reach a level of competency comparable with state-of-the-art deep-learning techniques on \gls{ale} games, for a fraction of their computational and storage cost.
Compared to state-of-the-art techniques, \glspl{tpg} reach comparable competency with one to three orders of magnitude less computations, and two to ten orders of magnitude less memory needed to store their inference model.
Recently, an extension of the \gls{tpg} model supporting continuous action space was proposed in order to target new learning environments, like time-series predictions~\cite{Kelly_Modular_2020}.

\textbf{Implementations of learning frameworks for \gls{tpg}}, coded in C++, Java and Python, can be found in open-source repositories.
The main motivations behind the creation of the \gls{gegelati} library is the desire to have an efficient, embeddable, portable, parallel and deterministic library.
Because of the efficiency and embeddability objectives, C++ was a natural choice for the development of \gls{gegelati}.
Previous open-source C++ implementations, including the reference C++ code from Kelly~\cite{Kelly_Emergent_2017}, were neither parallel nor deterministic.
The creation of a new library from scratch was further motivated by the low code quality of existing C++ implementation, notably due to a lack of code documentation and a monolithic code.

The purpose of this paper is not to advocate the learning efficiency of \glspl{tpg} against other machine learning techniques, which was already done in~\cite{Kelly_Emergent_2017, Kelly_Scaling_2018}.
Instead, this paper intends to present the original contributions for creating a customizable, scalable and deterministic implementation of \glspl{tpg}.

\section{\textsc{Gegelati}: Parallel, Efficient and Embeddable Library for TPGs}
\label{sec:gegelati}

\gls{gegelati} is an open-source framework, developed as a C++ library, for training and executing \glspl{tpg}.
From its inception, the \gls{gegelati} library has been conceived to foster its adaptability to diverse learning environments, and its portability to various architectures, without sacrificing its performance.
To this purpose, two original contributions have be integrated to the library: the parallelization of the deterministic learning process, presented in Section~\ref{sec:parallelism}; and the support for customizable \textit{instructions}, detailed in Section~\ref{sec:custom_instructions}.
An overview of additional features of the library is presented in Section~\ref{sec:features}.

\subsection{Deterministic Parallelism and Portability}
\label{sec:parallelism}

\textit{What are the motivations?}
Portability of the \gls{gegelati} library enables using it both on general-purpose and embedded architectures.
Indeed, when training a learning agent intended to run on an embedded system, a common design process is to prototype the agent first on a general-purpose processor before embedding it on the embedded target.
The portability also makes it possible to train a learning agent offline on a high-performance computing architecture, before deploying it on a less performing architecture for inference.

Parallelism of the learning process is an essential feature to accelerate the training of new learning agents, which fosters the adoption of new machine learning techniques.
Indeed, the breakthrough of deep-learning models is largely due to the acceleration of their training process with \glspl{gpu}~\cite{Krizhevsky_ImageNet_2012}.
Support for parallel computations is useful for general-purpose and high-performance computing architectures, but also for embedded systems which nowadays widely integrate heterogeneous \glspl{mpsoc}.

Determinism of a learning process is the property that ensures that given a set of initial conditions, the learning process will always end with the same result.
Determinism can only be obtained under the assumption that the state of the learning environment is itself changing deterministically, solely depending on the sequence of actions applied to it.
Determinism is a key feature, especially for a pseudo-stochastic learning process such as the training of \glspl{tpg}.
Indeed, the result of training may partially depend on luck, which is exactly why being able to deterministically reproduce a result is crucial.

The determinism is antagonistic with the parallelism and portability objectives, and with the stochastic nature of the learning process, which makes all these objectives challenging to implement jointly.
Indeed, parallelism is by nature a source of non-determinism as the simultaneity of computations accessing and modifying shared resources, often in an unknown order, tends to produce variable results.

\textit{How does the deterministic and scalable parallelism work?}
During the learning process of \glspl{tpg}, the most compute-intensive parts are the fitness evaluation of the \textit{policies}, and the mutations of the \textit{programs} added during the evolution process.
The fitness evaluation of individual \textit{policies} can be deterministically executed in parallel, on the conditions that: 1/ the learning environment can be cloned to evaluate several policies concurrently, and 2/ any stochastic evolution of the learning environment state can be controlled deterministically.
Under these conditions, the parallel evaluation of \textit{policies} is possible, as the topology of the \gls{tpg}, which is a shared resource for all \textit{policies}, is fixed during this evaluation process.
Similarly, the mutation of \textit{programs} can be applied deterministically in parallel.
Two kinds of mutations are applied to the \gls{tpg}: mutations affecting the graph topology by inserting new root \textit{teams} and edges; and mutations affecting \textit{instructions} of the \textit{programs} associated with the new edges.
While mutating the graph topology cannot be done in parallel, the graph being a shared resource, individual \textit{programs} are independent from each other and can be mutated in parallel.

To control a stochastic process, a \gls{prng} must be used each time a random number is needed.
Given an initial seed, a \gls{prng} produces a deterministic sequence of numbers.
To ensure full determinacy of the training of a \gls{tpg}, a unique \gls{prng} should be called in a fixed order during the whole training. 
Letting the parallel parts of the training process call the \gls{prng} directly is not possible, as the absolute order in which parallel computations occur is itself stochastic.
It is also not possible to give a pre-computed list of pseudo-random numbers to each parallel task, as the number of random numbers needed for each task is itself stochastic.
For example, when mutating a \textit{program}, mutations are applied iteratively until the program behavior becomes ``original'' compared to pre-existing \textit{programs} in the \gls{tpg}.
Hence, giving a fixed number of pre-computed random numbers for the \textit{program} mutations is not feasible.

The parallelization strategy adopted in \gls{gegelati} is based on the master/worker principle, with a distributed \gls{prng}. 
The principle of the distributed \gls{prng} is the use of two distinct \gls{prng} instances: the $\mathit{prng_{master}}$ and the $\mathit{prng_{worker}}$.
The $\mathit{prng_{master}}$ is exclusively used in the sequential parts of the learning process, which confers a deterministic nature to its usage, given an initial seed.
Besides being used for stochastic tasks performed sequentially, like \gls{tpg} topology mutations for example, the $\mathit{prng_{master}}$ is also used to generate a seed for each parallel worker task.
In each worker task, a private $\mathit{prng_{worker}}$ is instantiated, and initialized with the seed provided by the $\mathit{prng_{master}}$.
Since all calls to the \gls{prng} from the worker tasks exclusively use their private $\mathit{prng_{worker}}$, the random number sequences generated in each parallel task are deterministic.

\begin{figure}[t]
	\begin{minipage}[t]{0.44\linewidth}
		\vspace{0pt} 
		\begin{procedure}[H]
			\caption{EvaluateAllPolicies()}
			\label{proc:master}
			\KwIn{\hspace{0.4em} \gls{tpg}: $G = \langle \mathit{Teams}, \mathit{Edges}\rangle$}
			\KwData{\hspace{0.85em} \gls{prng}: $\mathit{prng_{master}}$ \\
				\hspace{3.45em} Job queue: $\mathit{JobQ}$ \\ 
				\hspace{3.45em} Result Queue: $\mathit{ResultQ}$}
			\tcc{Prepare jobs}
			idx = 0 \\
			\ForEach{root $\in$ G.Teams}{
				seed = $\mathit{prng_{master}}$.getNumber() \\
				job = \{ idx++, seed, root \} \\
				jobQ.push(job)
			} 
			\tcc{Start parallel threads}
			\For{i = 1 \KwTo $\mathit{Num}_{\mathit{PE}} - 1$}{
				Spawn thread: Worker(G, JobQ, ResultQ)
			}
			Call Worker(G, JobQ, resultQ) \\
			Join all threads \\
			\tcc{Post-Process Results and Trace} 
			Sort ResultQ in result.jobId order\\
			\ForEach{result $\in$ resultQ}{
				Post-process result.trace // Archiving \cite{Kelly_Emergent_2017} \\
				...
			}
		\end{procedure}
	\end{minipage}
	\hspace{2em}
	\begin{minipage}[t]{0.44\linewidth}
		\begin{procedure}[H]
			\caption{Worker()}
			\label{proc:worker}
			\KwIn{\hspace{0.6em}\gls{tpg}: $G = \langle \mathit{Teams}, \mathit{Edges}\rangle$ \\
				\hspace{3.4em} Job queue: $\mathit{JobQ}$ \\
				\hspace{3.4em} Result queue: $\mathit{ResultQ}$}
			\KwData{\hspace{0.65em} \gls{prng}: $\mathit{prng_{worker}}$ \\
				\hspace{3.35em} Learning environment twin: $LE$ }
			\tcc{Poll for job}
			\While{JobQ.hasJob()}{
				\tcc{Setup for policy evaluation}
				job =  jobQ.getNextJob() \\
				root = job.root \\
				$\mathit{prng_{worker}}$.reset(job.seed) \\
				LE.reset($\mathit{prng_{worker}}$.getNumber()) \\
				\tcc{Evaluate policy fitness}
				trace = evaluate(G, root, LE, $prng_{worker}$) \\
				result.jobId = job.id \\
				result.trace = trace \\
				resultQ.push(result)   	    	
			}
		\end{procedure}
	\end{minipage}
\vspace{-1.5em}
\end{figure}

The pseudo-code of the master and worker tasks for the policy fitness evaluation are presented in Procedures~\ref{proc:master} and~\ref{proc:worker}, respectively.
Communications between the tasks and load balancing of the computations are supported by a job queuing mechanism based on two queues: \textit{JobQ} and \textit{ResultQ}.
Each policy evaluation job, prepared by the master procedure, encapsulates a unique job identifier \texttt{id}, a seed provided by the $\mathit{prng_{master}}$, and a root \textit{team} from the \gls{tpg}.
All jobs are pushed in the \textit{JobQ} queue before spawning as many worker threads as the number of secondary \glspl{pe} in the target architecture.
For each job it acquires from the \textit{jobQ} queue, the worker procedure resets its $\mathit{prng_{worker}}$ using the seed contained in the job.
Before evaluating the fitness of the root \textit{team} contained in a job, the worker procedure resets its private copy of the learning environment, using a number given by the $\mathit{prng_{worker}}$.
As a result of the policy fitness evaluation, described in details in~\cite{Kelly_Emergent_2017}, a result object encapsulating execution traces for the job is pushed in the \textit{resultQ}.
When all jobs have been processed, and all workers terminated, the master procedure is responsible for post-processing the traces stored in the \textit{resultQ}.
To ensure determinism of this post-processing, results stored in the \textit{resultQ} are first sorted in ascending \texttt{job.id} order.

The master and worker procedures used for parallelizing the mutations of \textit{programs} are similar to the one used for policy fitness evaluation, with the difference that jobs encapsulate \textit{programs} instead of root \textit{teams}.
In Section~\ref{sec:experiments}, experiments on three different multicore architectures show the scalability and load balancing capability of the proposed deterministic and parallel \gls{tpg} implementation.

\subsection{Customizable Instruction Set}
\label{sec:custom_instructions}
\textit{What are the motivations?} 
In the seminal work on \glspl{tpg}~\cite{Kelly_Emergent_2017}, the \textit{instructions} used in the \textit{programs} are chosen exclusively among the following eight instructions: 4 binary operators $\{+,-,\times, \div\}$, 3 mathematical functions $\{cos, ln, exp\}$, and 1 conditional statement $res \leftarrow (a < b)? -a : a$.
To further simplify the execution and mutation of \textit{programs}, it was assumed that instructions only handle \texttt{double} operands.

As shown in related genetic programming works~\cite{Atkins_domain_2011, Real_AutoML_2020}, using a broader set of instructions with diverse data types can help improve the performance of learning agents, at the cost of longer training time.
The extension of the instruction set used in the \textit{programs} of the \glspl{tpg} has already been proposed in~\cite{Kelly_Temporal_2020}, where a set of instructions for 2D images operands is added, and in~\cite{Gesny_CBWAR_2018}, with instructions accepting thirteen operands tailored for predicting properties of the learning environment.
In \gls{gegelati}, both the number and types of operands, and the nature of instructions used in \textit{programs} can be fully customized.
Besides making the training more efficient for specific learning environments, this customization feature may also be used to increase the efficiency of the \gls{tpg} execution on specific hardware.
Indeed, using an instruction set mirroring the instruction set of the architecture used for its execution may help increase the speed and the power efficiency of the \gls{tpg} execution.

\textit{How are customizable instructions supported?}
\begin{figure}[b]
	\begin{minipage}[t]{0.4\linewidth}
		\vspace{0pt} 
		\begin{tikzpicture}[]
		\node(O) at (1.8,4.25){		
			\tikz{
				\node(classH) at (0,1.0){
					\tikz {\draw[thick,color=black] (0,0) rectangle ++(3.6,0.5);} 
				};
				\node[at={($(classH.center)$)},anchor=center] {Instruction};   
				\node(classD) at (0,0.5){
					\tikz {\draw[thick,color=black] (0,0) rectangle ++(3.6,0.5);} 
				};
				\node[at={($(classD.west)+(0.05,0)$)},anchor=west] {\small\texttt{+operandTypes[]}};   
				\node(classF) at (0,0.0){
					\tikz {\draw[thick,color=black] (0,0) rectangle ++(3.6,0.5);} 
				};
				\node[at={($(classF.west)+(0.05,0)$)},anchor=west] {  \small\texttt{+execute(operands[])}};     
			}
		};			
		\node(I) at (1.8,2.0){		
			\tikz{
				\node(classH) at (0,1.5){
					\tikz {\draw[thick,color=black] (0,0) rectangle ++(3.6,0.5);} 
				};
				\node[at={($(classH.center)$)},anchor=center] {ProgramLine};   
				\node(classD) at (0,0.65){
					\tikz {\draw[thick,color=black] (0,0) rectangle ++(3.6,1.2);} 
				};
				\node[at={($(classD.west)+(0.05,0)$)},anchor=west, text width=2.9] {\small\texttt{+instruction*} \small\texttt{+operandAdresses[]}
					\small\texttt{+register}};   
				\node(classF) at (0,-0.2){
					\tikz {\draw[thick,color=black] (0,0) rectangle ++(3.6,0.5);} 
				};
			}
		};
		%
		\node(D) at (1.8,-0.75){		
			\tikz{
				\node(dataH) at (0,1.2){
					\tikz {\draw[thick,color=black] (0,0) rectangle ++(3.6,0.5);} 
				};
				\node[at={($(dataH.center)$)},anchor=center] {DataSource};   
				\node(dataD) at (0,0.7){
					\tikz {\draw[thick,color=black] (0,0) rectangle ++(3.6,0.5);} 
				};
				\node[at={($(dataD.west)+(0.05,0)$)},anchor=west] {\small\texttt{-data}};   
				\node(dataF) at (0,-0.15){
					\tikz {\draw[thick,color=black] (0,0) rectangle ++(3.6,1.2);} 
				};
				\node[at={($(dataF.west)+(0.05,0)$)},anchor=west, text width=3.6] {  \small\texttt{+canProvide(address,type)}
					\small\texttt{+getData(address,type)}
					\small\texttt{+setData(address,data)}};     
			}
		};
		\node(capt) at (1.2,-2.5){
			\begin{minipage}[t]{4.7cm}
			\captionof{figure}{Class diagrams of the data structures for customizable instructions.}
			\label{fig:structures}
			\end{minipage}
		};
		\end{tikzpicture}
	\end{minipage}
	\begin{minipage}[t]{0.5\linewidth}
		\begin{procedure}[H]
			\caption{ExecuteProgram()}
			\label{proc:execute}
			\KwIn{Program: \textit{p} \\
				\hspace{3em}Data sources: \textit{data}}
			\ForEach{line $\in$ $p$}{
				instruction = \textit{line}.instruction \\
				operands[] = \{ $\emptyset$ \} \\ 
				nbOperands = instruction.operandTypes.size() \\
				\For{$i=0$ \KwTo nbOperands-1}{
					\label{line:op_begin}
					type = instruction.dataTypes[i] \\
					address = \textit{line}.operandAdress[i] \\
					\eIf{\textit{data}.canProvide(type, address)}{
						operand = data.getData(type, address) \\
						operands.insert(operand)
					}{
						Exit with an error
					}						
				} 
			    \label{line:op_end}
				result = instruction.execute(operands) \\
				data.set(line.register,result)
			}
		\end{procedure}
	\end{minipage}
\end{figure}
%
The support for customizable instructions within \gls{gegelati} is based on the three classes presented in Figure~\ref{fig:structures}.
When creating a new training environment, a developer may create her own set of instructions, by creating new classes inheriting from the \texttt{Instruction} class.
With the \texttt{operandTypes} attribute, each instruction declares the number and type of operands it accepts when calling its \texttt{execute()} method.
Currently, to keep the management of registers simple during program execution, only \texttt{double} results can be produced by the \texttt{execute()} method.
Each \textit{line} of a \textit{program} references an \texttt{instruction} from the set of available instructions, a destination \texttt{register} to store the result of its execution, and the addresses of operands to process, selected among all available data sources.
The data sources accessible to the \textit{lines} comprise both the registers used for storing instruction results, and the state of the learning environment.
Data sources classes must inherit from the \texttt{DataSource} class which acts as a wrapper between the data and the \textit{program} execution engine.

Procedure~\ref{proc:execute} presents the simplified pseudo-code for executing a \textit{program} modeled with the classes from Figure~\ref{fig:structures}.
The core of the mechanism supporting customizable instructions lies between lines~\ref{line:op_begin} and~\ref{line:op_end} of Procedure~\ref{proc:execute}.
For each operand of each \textit{line}, the algorithm checks whether the data sources can provide the requested operand type at the requested address.
If the data type can be provided by the data sources at the requested address, the data is fetched from the data sources, and later used for executing the \texttt{instruction} of the current \texttt{line} of the \textit{program}.
Otherwise, the program execution is terminated, which does not occur in practice, as the operand data types are taken into consideration when performing \textit{program} mutations in \gls{gegelati}.
It is important to note that the \texttt{getData()} method may return data whose type differs from the native data type stored within the data source.
For example, a data source storing screen pixels as \texttt{char} values can automatically return an equivalent \texttt{double} value, or even a neighborhood of 3-by-3 pixels when an operand of type \texttt{char[3][3]} is requested. 

\setlength{\intextsep}{0pt}
\begin{wrapfigure}[4]{r}{8.6cm}
\begin{lstlisting}[style=customcpp, caption=\texttt{LambdaInstruction} usage example, label=lst:lambda]
auto myInstruction = LambdaInstruction<int, char[2]>(
  [](int a, char[2] b)->double {return a*(b[0] + b[1]);});
\end{lstlisting}
\end{wrapfigure}
To ease the creation of new \textit{instructions} for each training environment, a utility class \texttt{LambdaInstruction} is proposed in \gls{gegelati}.
The template class \texttt{LambdaInstruction} supports the creation of instructions for any number of operands, and for operands with primitive and non-primitive types as well as 1D and 2D C-Style arrays.
A code snippet illustrating the creation of an instruction with the \texttt{LambdaInstruction}  class is given in Listing~\ref{lst:lambda}.
In this example, an instruction taking an \texttt{int} operand, and a 1D array of \texttt{char} is declared, using a simple C++ lambda function.

%
\subsection{Library Features}
\label{sec:features}
In addition to the two contributions detailed in previous sections, the \gls{gegelati} library showcases several important features for easing its integration in new projects.

\textit{Configuration and logging support.} 
Interfacing a machine learning library with third-party tools, notably scripting and data analysis tools, is an indispensable feature.
To this purpose, \gls{gegelati} supports the import and export of information in several standard file formats.
A \texttt{json} importer is available for configuring the meta parameters of the \gls{tpg} training process. 
This \texttt{json} importer eases the exploration of many meta parameter configurations from a simple script, without having to re-build the library code.
A \gls{tpg} exporter in the \texttt{dot} format can be used to visualize the topology of the graph along the training process.
The exported \texttt{dot} files also embed all the information needed to import a pre-trained \gls{tpg}.
A logging system exporting training statistics in the \texttt{csv} format is also implemented in \gls{gegelati}.
Through a simple class specialization mechanism, this logging system may be extended to export information specific to a user's learning environment.

\textit{Code quality.}
Maintaining a high quality of code is imperative to foster the maintainability, the evolutivity, and the trustworthiness in a library.
Strict coding rules are enforced in \gls{gegelati}, notably by automatically checking that all data structures and functions are properly commented, and causing a build failure otherwise.
Publicly hosted continuous integration is used to ensure that code coverage of the unit tests cover 100\% of the library code~\cite{GEGELATI_Git}.
This continuous integration also checks the portability of the library on Linux with gcc and clang, on Windows with MSVC19, on and macOS with clang.
Public continuous code quality analysis is also used to enforce the reliability, the security, and the maintainability of the code~\cite{GEGELATI_Git}.

\textit{Extension-friendly code modularity.}
Adopting a modular and hierarchical code organization of the code favors its evolution and its extension, either by replacing existing modules, or by adjuncting new ones.
In \gls{gegelati}, separate modules with as few inter-dependencies as possible have been set up for each concept in the code.
For example, classes modeling the topology of a \gls{tpg} do not depend on the code responsible for mutating it, or the code for executing it.
This separation of concern can be exploited to create specialization of a given module for a specific purpose.
For example, while the default learning algorithm assumes a classic reinforcement learning setup with one agent interacting with its environment, an alternative algorithm has been implemented to train two or more agents to play competitive games.
Another alternative learning algorithm is also available for learning environments representing a classification problem.

\textit{Open source applications.}
The public availability of projects based on a library is a way to demonstrate in a reproducible way the capabilities of this library.
Existing projects can also serve as examples or even as a base, for a developer starting a new project with a library.
The public repository of \gls{gegelati}~\cite{GEGELATI_Git} hosts several projects with diverse learning environments, including: a wrapper for the \acrfull{ale}~\cite{Bellemare_Arcade_2013}, simple 2-player tic-tac-toe and Nim games, a physical robotic arm control and an inverted pendulum environments, and a MNIST~\cite{LeCun_Gradient_1998} classification environment.

\section{Experiments}
\label{sec:experiments}
The \gls{gegelati} implementation of the \gls{tpg} learning process is evaluated in different scenarios to assess its capabilities and performance.
Version \texttt{0.4.0} of the \gls{gegelati} was used in all experiments.
In Section~\ref{sec:expe_performance}, the rapidity and the scalability of the training process is evaluated on several multicore architectures.
In Section~\ref{sec:expe_instruction}, the potential of customizable instructions is demonstrated on an inverted pendulum environment.

\subsection{Sequential Performance and Scalability}
\label{sec:expe_performance}

\textit{Sequential.}
The implementation of the contributions presented in Section~\ref{sec:gegelati} incurs an additional complexity in the code of \gls{gegelati} compared to the reference \gls{tpg} implementation by Stephen Kelly~\cite{Kelly_Emergent_2017,Kelly_Scaling_2018}.
To assess the efficiency of \gls{gegelati} compared to this reference implementation, the two codes have been used to train \glspl{tpg} in the \gls{ale} environment.
The meta-parameters values defined in~\cite{Kelly_Scaling_2018} have been used in all experiments, except the number of games played per evaluation, which was set to 1 to accelerate the experiments.
\glspl{tpg} were trained for 50 generations on 5 games with diverse complexities: \texttt{alien}, \texttt{asteroids}, \texttt{centipede}, \texttt{fishing\_derby}, and \texttt{frostbite}.
Due to the non-deterministic stochastic nature of the Atari 2600 emulator powering the \gls{ale}~\cite{Bellemare_Arcade_2013}, 5 runs were executed for each game.
This experiment was executed on one core of an Intel Xeon CPU E5-2690 processor, running with Linux.

Training time of the \glspl{tpg} measured in this experiment are plotted in Figure~\ref{fig:sequential}.
For each game and implementation, the plot illustrates the minimum, maximum, and average training time, relatively to the average training time of the reference implementation.
The absolute training time, in minutes, of the reference implementation is printed in the plot. 

\definecolor{korange}{RGB}{247,150,70}
\definecolor{kpink}{RGB}{0,120,181}
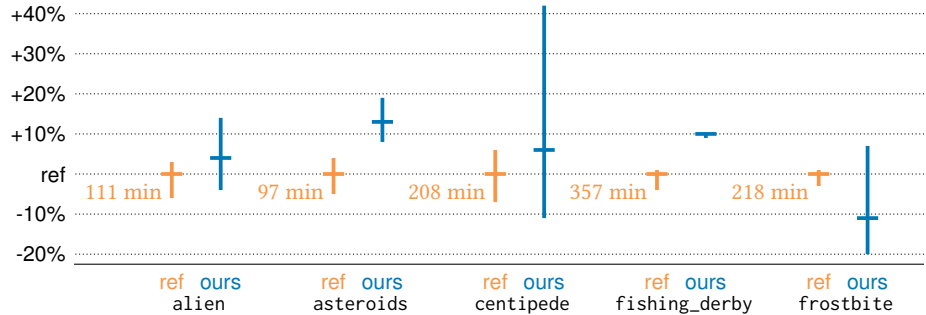
\begin{figure}
	\centering
	\begin{tikzpicture}[line join=round,>=triangle
	45,x=1.0cm,y=0.8cm]
	
	\def\games{{"alien","asteroids","centipede","fishing\_derby","frostbite"}}
	\def\results{{
		{-6,0,3,-4,4,14,111},
		{-5,0,4, 8,13,19,97},
		{-7,0,6,-11,6,42,208},
		{-4,0,1,9,10,10,357},
		{-3,0,1,-20,-11,7,218},
	}}
	\def\textHeight{6}
	\def\yScale{15}
	\def\yOffset{1.5}
	\def\spacing{2.15}
	\def\sepRefOur{0.65}
	\draw[-,color=black] (-1.3,0) -- (10,0);
	\draw[densely dotted,color=black] (10,\yOffset) -- (-1.3,\yOffset) node[left]{\labelFont{ref}};
	\foreach \i in {-20,-10,+10,+20,+30,+40}
    \draw[densely dotted,color=black] (10,\yOffset+\i/\yScale) -- (-1.3,\yOffset+\i/\yScale) node[left]{\labelFont{\i\%}};
	
	\foreach \x in {0,...,4}{
	\draw[shift={(\spacing*\x,0)},color=korange] node[below, text height=\textHeight]{\labelFont{ref}};
	\draw[shift={(\spacing*\x+\sepRefOur,0)},color=kpink] node[below, text height=\textHeight]{\labelFont{ours}};
	\draw[shift={(\spacing*\x+0.36,-0.35)},color=black] node[below, text height=\textHeight]{\labelFont{\pgfmathparse{\games[\x]}\texttt{\pgfmathresult}}};
	\pgfmathtruncatemacro{\kmin}{\results[\x][0]}
	\pgfmathtruncatemacro{\kavg}{\results[\x][1]}
	\pgfmathtruncatemacro{\kmax}{\results[\x][2]}
	\pgfmathtruncatemacro{\kabs}{\results[\x][6]}
	\draw[shift={(\spacing*\x,0)},color=korange,line width=0.50mm] (0,\kmin/\yScale+\yOffset) -- (0,\kmax/\yScale+\yOffset);
	\draw[shift={(\spacing*\x,0)},color=korange,line width=0.50mm] (-4pt,\yOffset) -- (4pt,\yOffset);
	\draw[shift={(\spacing*\x,0)},color=korange,line width=0.50mm] (0,\yOffset) node[below left]{\kabs{}~min};
	\pgfmathtruncatemacro{\kmin}{\results[\x][3]}
	\pgfmathtruncatemacro{\kavg}{\results[\x][4]}
	\pgfmathtruncatemacro{\kmax}{\results[\x][5]}
	\draw[shift={(\spacing*\x+\sepRefOur,0)},color=kpink,line width=0.50mm] (0,\kmin/\yScale+\yOffset) -- (0,\kmax/\yScale+\yOffset);
	\draw[shift={(\spacing*\x+\sepRefOur,0)},color=kpink,line width=0.50mm] (-4pt,\kavg/\yScale+\yOffset) -- (4pt,\kavg/\yScale+\yOffset);
    }
	\end{tikzpicture}
		\vspace{-1em}
    \caption{Relative sequential training time of \gls{gegelati} (ours) and Kelly's code (ref)~\cite{Kelly_Emergent_2017}. On each game, \glspl{tpg} were trained 5 times, for 50 generations. For each game and code couple, the vertical line spans from the minimum to the maximum training times, and the horizontal line is the average training time. All times are relative to the average ref training time printed in the plot.}
    \label{fig:sequential}
    \vspace{-1.5em}
\end{figure}  

As can be seen in the experimental results in Figure~\ref{fig:sequential}, \gls{gegelati} is on average only 4.4\% slower than the reference code.
In the best case, for the \texttt{frostbite} game, \gls{gegelati} is on average 11\% faster than the reference code, while it is on average 13\% slower in the worst case, for the \texttt{asteroids} game.
In these experiments, the variability of the training times, revealed by the min-max ranges, seems to be more important with \gls{gegelati}, which indicates a greater sensibility to variations of the \gls{tpg} topology induced by the training process.
The stochastic nature of the learning environment impacts the fitness of the trained agent, the chances of reaching a game-over, and the topology of the \gls{tpg}, which are as many variables impacting the training time.
Due to these numerous variables, an in-depth analysis of these results is out of the scope of this paper.
Nevertheless, a fair conclusion to this experiment is that, despite the overhead of its additional mechanisms for supporting deterministic parallelism and customizable instructions, the sequential performance of \gls{gegelati} are only slightly lower than the reference code. 

\textit{Parallel scalability.}
To assess the efficiency, the portability, and the scalability of the parallelization strategy implemented in \gls{gegelati}, experiments were conducted on 3 distinct architectures: 1/ a high-performance Intel Xeon CPU E5-2690 processor with 24 cores, 2/ an Intel i7-8650U CPU with 4 cores, and 3/ a Samsung Exynos5422, an embedded heterogeneous chip with four \textsc{Arm} A15 cores, and four A7 cores.
To assess the scalability of the parallelism on each target, the training time was measured for a number of thread varying between 1 and the number of physical core of the architecture.
No affinity constraint was set on the threads, thus letting the Linux operating system handle the scheduling.
On the Exynos chip, the \texttt{taskset} command was used to control the number of cores of each type, A15 and A7, available to the threads.
For each number of threads, and for each game on the Intel CPUs, the experiment was repeated 5 times with different seeds.

On the Xeon and i7 CPUs, the \gls{ale} learning environment was used, with the same 5 games and meta-parameters as in the sequential experiments.
On the Exynos CPU, a more lightweight learning environment, the inverted pendulum learning environment was used as a more realistic use case for an embedded processor.
For each number of threads, the experiment was conducted 5 times with different seeds.
On the pendulum learning environment, full determinism of the training process is observed, in all configurations.
\vspace{1em}

\begin{figure}[ht]
	\subcaptionbox{Intel Xeon CPU (24 cores)}{
	\centering
	\begin{tikzpicture}[x=0.20cm,y=0.12cm,line join=round,>={Triangle[angle=45:7pt]}]
	
	\def\xScale{1}
	\def\yScale{1}
	\def\xMax{28}
	\draw[->, color=black] (1*\xScale,0) -- (\xMax*\xScale,0) node[below]{$nb_{cores}$};
	\foreach \x in {1,6,12,18,24}{
	\draw[color=black] (\x*\xScale,-2pt) -- (\x*\xScale, 2pt) node[below,text height=8]{\x};
	\draw[densely dotted, color=black] (\x*\xScale,0) -- (\x*\xScale, 24*\yScale);
    }

	\draw[->, color=black] (1*\xScale,0) -- (1*\xScale,26*\yScale) node[below right]{$Speedup$};
	\foreach \y in {5,10,15,20}
	\draw[densely dotted,color=black] (26*\xScale, \y*\yScale) -- (1*\xScale,\y*\yScale) node[left]{x\y};
	
	\draw[color=kpink,thick] plot[mark=*,mark size=0pt] file {avgSpeedup24.txt} node [right] {};
	\draw[densely dotted,color=korange,thick] plot[mark=*,mark size=0pt] file {minSpeedup24.txt} node [right] {};
	\draw[densely dotted,color=korange,thick] plot[mark=*,mark size=0pt] file {maxSpeedup24.txt} node [right] {};
	
	\end{tikzpicture}
	\label{fig:speedup_xeon}
}
\hspace{1em}
\subcaptionbox{Intel i7 CPU (4 cores)}{
	\centering
	\begin{tikzpicture}[x=1.4cm,y=0.6cm,line join=round,>={Triangle[angle=45:7pt]}]
	
	\def\xScale{1}
	\def\yScale{1}
	\def\xMax{4.6}
	\draw[->, color=black] (1*\xScale,0) -- (\xMax*\xScale,0) node[below]{$nb_{cores}$};
	\foreach \x in {1,2,3,4}{
		\draw[color=black] (\x*\xScale,-2pt) -- (\x*\xScale, 2pt) node[below,text height=8]{\x};
		\draw[densely dotted, color=black] (\x*\xScale,0) -- (\x*\xScale, 4.5*\yScale);
	}
	
	\draw[->, color=black] (1*\xScale,0) -- (1*\xScale,5*\yScale) node[below right]{$Speedup$};
	\foreach \y in {1,2,3,4}
	\draw[densely dotted,color=black] (4.3*\xScale, \y*\yScale) -- (1*\xScale,\y*\yScale) node[left]{x\y};
	
	\draw[color=kpink,thick] plot[mark=*,mark size=0pt] file {avgSpeedup4.txt} node [right] {};
	\draw[densely dotted,color=korange,thick] plot[mark=*,mark size=0pt] file {minSpeedup4.txt} node [right] {};
	\draw[densely dotted,color=korange,thick] plot[mark=*,mark size=0pt] file {maxSpeedup4.txt} node [right] {};	
	\end{tikzpicture}
	\label{fig:speedup_i7}
}
\vspace{-1em}
\caption{Training Scalability on Homogeneous Multicore CPUs}
\label{fig:speedup}
\vspace{-1.5em}
\end{figure}
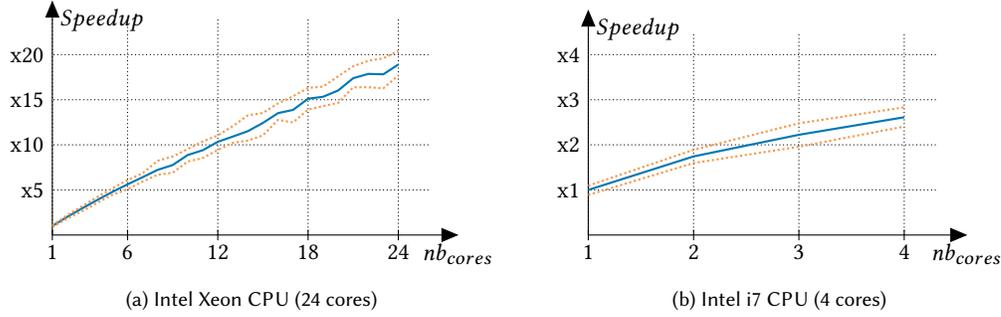

\begin{wrapfigure}[24]{r}{6.2cm}
	\begin{tikzpicture}[x=0.64cm,y=0.6cm,line join=round,>={Triangle[angle=45:7pt]}]
	\def\xScale{1}
	\def\yScale{1}
	\def\xMax{9}
	\draw[->, color=black] (1*\xScale,0) -- (\xMax*\xScale,0) node[below]{$nb_{cores}$};
	\foreach \x in {1,2,3,4,5,6,7,8}{
		\draw[color=black] (\x*\xScale,-2pt) -- (\x*\xScale, 2pt) node[below,text height=8]{\x};
		\draw[densely dotted, color=black] (\x*\xScale,0) -- (\x*\xScale, 4.5*\yScale);
	}
	
	\draw[densely dotted,color=black] (4.5,0) -- (4.5,-1.2);
	\draw (2.5,-0.8) node[text height=8]{A15 cores};
	\draw (6.5,-0.8) node[text height=8]{A7 cores};
	
	\draw[->, color=black] (1*\xScale,0) -- (1*\xScale,5*\yScale) node[below right]{$Speedup$};
	\foreach \y in {1,2,3,4}
	\draw[densely dotted,color=black] (8.3*\xScale, \y*\yScale) -- (1*\xScale,\y*\yScale) node[left]{x\y};
	
	\draw[color=kpink,thick] plot[mark=*,mark size=0pt] file {avgSpeedupOdroid.txt} node [right] {};
	\draw[densely dotted,color=korange,thick] plot[mark=*,mark size=0pt] file {minSpeedupOdroid.txt} node [right] {};
	\draw[densely dotted,color=korange,thick] plot[mark=*,mark size=0pt] file {maxSpeedupOdroid.txt} node [right] {};
	
\end{tikzpicture}
	\caption{Scalability on Exynos5422 (8 big.LITTLE cores)}
	\label{fig:speedup_odroid}
\vspace{1em}
\begin{tikzpicture}[x=0.011cm,y=0.00035cm,line join=round,>={Triangle[angle=45:7pt]}]
	\def\xScale{1}
	\def\yScale{1}
	\def\xMax{470}
	\def\yMax{10000}
	\draw[->, color=black] (1*\xScale,0) -- (\xMax*\xScale,0) node[below]{$Gen.$~\hspace*{0.8em}};
	\foreach \x in {100,200,300,400}{
		\draw[color=black] (\x*\xScale,-2pt) -- (\x*\xScale, 2pt) node[below,text height=8]{\x};
		\draw[densely dotted, color=black] (\x*\xScale,0) -- (\x*\xScale, 4.5*\yScale);
	}
	
	\draw[->, color=black] (0,0) -- (0,\yMax*\yScale) node[below left]{$Score$};
	\foreach \y in {2000,4000,6000,8000}
	\draw[color=black] (2pt, \y*\yScale) -- (-2pt,\y*\yScale) node[left]{\y};
	
	\draw[opacity=0,fill=kpink,fill opacity=0.2] plot[mark=*,mark size=0pt] file {withInstructionsMinMax.txt} node [right] {};
	\draw[opacity=0,fill=korange,fill opacity=0.2] plot[mark=*,mark size=0pt] file {withoutInstructionsMinMax.txt} node [right] {};
	
	\draw[color=kpink] (340,8400) node {\smallFont{\textit{ISet}$_{\mathit{complex}}$}};
	\draw[color=kpink,thick] plot[mark=*,mark size=0pt] file {withInstructionsAvg.txt} node [right] {};
	\draw[densely dotted,color=kpink] plot[mark=*,mark size=0pt] file {withInstructionsMax.txt} node [right] {};
	\draw[densely dotted,color=kpink] plot[mark=*,mark size=0pt] file {withInstructionsMin.txt} node [right] {};
	
	\draw[color=korange] (340,4900) node {\smallFont{\textit{ISet}$_{\mathit{simple}}$}};
	\draw[color=korange,thick] plot[mark=*,mark size=0pt] file {withoutInstructionsAvg.txt} node [right] {};
	\draw[color=korange, densely dotted] plot[mark=*,mark size=0pt] file {withoutInstructionsMax.txt} node [right] {};
	\draw[color=korange, densely dotted] plot[mark=*,mark size=0pt] file {withoutInstructionsMin.txt} node [right] {};
	\end{tikzpicture}
	\vspace*{-2.5em}
	\caption{Champion policy score on \texttt{frostbite} over 400 generations with 2 instruction sets. For each set, the thick line represents the champion policy score averaged on 5 trainings. The colored background covers the area from the minimum to the maximum champions score among the 5 trainings.}
	\label{fig:custom_iset}
\end{wrapfigure}
%
Figure~\ref{fig:speedup} presents the results of the scalability evaluation of the training process of \gls{gegelati} on the homogeneous multicore CPUs. 
For each target architecture, the speedup compared to the sequential training time is plotted for a varying number of threads running the worker process.
The thick line represents the average speedup observed for all configurations, and the dotted lines represent the minimum and maximum speedup observed.
As can be seen in these results, the parallelization of the learning process is highly scalable, with an average speedup of 18.9 on the 24 cores of the Xeon CPU, and a speedup of 2.61 on the 4 cores of the i7 CPU.

Figure~\ref{fig:speedup_odroid} presents the experimental results for the Exynos chip. 
On this heterogeneous chip, the average speedup is 4.12 when using all 8 cores of the architecture, compared to a single A15 core.
There is a clear benefit in using the four A7 cores, in addition to the four A15 cores, as the average speedup when using only the four A15 cores is only 2.32.

In experiments based on the \gls{ale} games, the training time is dominated by the computation time needed to evaluate the fitness of each policy by playing the game, which takes up 99.8\% of all training time.
On the pendulum learning environment, 40\% of the training time is spent performing the graph mutations.
In this environment, mutations are sped-up by a factor 3.80 on average, and policy evaluation by a factor 4.41, on the 8 heterogeneous cores of the Exynos chip.

\subsection{Customizable Instruction Sets}
\label{sec:expe_instruction}

To illustrate the utility of customizable instructions, \gls{gegelati} was trained for 400 generations on the \texttt{frostbite} game from the \gls{ale}, using two different sets of instructions, namely \textit{ISet}$_{\mathit{simple}}$ and \textit{ISet}$_{\mathit{complex}}$.
The \textit{ISet}$_{\mathit{complex}}$ set contains the eight instructions from Kelly's work: $\{+,-,\times, \div, cos, ln, exp, <\}$, and the \textit{ISet}$_{\mathit{complex}}$ contains only the following five low-complexity instructions $\{+,-,\times, \div, <\}$.
For each instruction set, the training was performed with 5 different seeds.

For each instruction set, Figure~\ref{fig:custom_iset} plots statistics of the score of the champion policies observed during training.
The thick lines represent the average score of the champion policies during the training process.
The surrounding colored zone covers the area between the minimum and maximum scores observed during the five trainings for each instruction set.
The trainings based on the \textit{ISet}$_{\mathit{complex}}$ instruction set outperform the trainings based on the \textit{ISet}$_{\mathit{simple}}$ set.
On average, the score of the champion policies with the \textit{ISet}$_{\mathit{simple}}$ set plateau at 4626 points, whereas the champion policies built with the \textit{ISet}$_{\mathit{complex}}$ set reach 8042 points.
The gap between the training efficiencies with the two instruction sets is also revealed by the distance between the best score obtained after 400 generations for the \textit{ISet}$_{\mathit{simple}}$ set: 5470 points, which is far below the worst score obtained with the \textit{ISet}$_{\mathit{complex}}$ set: 6450 points.
This experiment shows that by providing support for customizable instruction sets, \gls{gegelati} unlocks the possibility to tailor the training instruction set to maximize the training efficiency.

\section{Conclusion}
\label{sec:conclusion}
The \gls{gegelati} C++ library for the training of \glspl{ai} based on \glspl{tpg} was introduced in this paper.
Two original features of the library, namely its scalable and deterministic parallelism, and its customizable instructions, confer \gls{gegelati} great portability on various types of architectures.
The raw performance, the scalability, and the customizable efficiency of the training process of \glspl{tpg} was assessed experimentally using state-of-the-art reinforcement learning environments.
A direction for future work is the creation of a methodology for selecting appropriate instruction sets to maximize the efficiency of the training process, while jointly optimizing the time and energy efficiency of selected instructions.

\bibliographystyle{ACM-Reference-Format}
\bibliography{gegelati}

\end{document}